\documentclass[conference]{IEEEtran}
\IEEEoverridecommandlockouts
\usepackage{cite}
\usepackage{amsmath,amssymb,amsfonts}
\usepackage{algorithmic}
\usepackage{graphicx}
\usepackage{textcomp}
\usepackage{xcolor}
\usepackage{url}
\usepackage{caption}
\usepackage{subcaption}
\def\BibTeX{{\rm B\kern-.05em{\sc i\kern-.025em b}\kern-.08em
    T\kern-.1667em\lower.7ex\hbox{E}\kern-.125emX}}

\begin{document}

\title{Multi-output Deep-Supervised Classifier Chains for Plant Pathology
}

\author{\IEEEauthorblockN{Jianping Yao}
\IEEEauthorblockA{\textit{School of  Information and Communications Technology} \\
\textit{University of Tasmania}\\
Launceston, Australia \\
jianping.yao@utas.edu.au}
\and
\IEEEauthorblockN{Son N. Tran\IEEEauthorrefmark{1}\thanks{\IEEEauthorrefmark{1}Son N. Tran is the corresponding author.}}
\IEEEauthorblockA{\textit{School of  Information and Communications Technology} \\
\textit{University of Tasmania}\\
Launceston, Australia \\
sn.tran@utas.edu.au}}

\maketitle

\begin{abstract}
Plant leaf disease classification is an important task in smart agriculture which plays a critical role in sustainable production. Modern machine learning approaches have shown unprecedented potential in this classification task which offers an array of benefits including time saving and cost reduction. However, most recent approaches directly employ convolutional neural networks where the effect of the relationship between plant species and disease types on prediction performance is not properly studied. In this study, we proposed a new model named Multi-output Deep Supervised Classifier Chains (Mo-DsCC) which weaves the prediction of plant species and disease by chaining the output layers for the two labels. Mo-DsCC consists of three components: A modified VGG-16 network as the backbone, deep supervision training, and a stack of classification chains. To evaluate the advantages of our model, we perform intensive experiments on two benchmark datasets Plant Village and PlantDoc. Comparison to recent approaches, including multi-model, multi-label (Power-set), multi-output and multi-task, demonstrates that Mo-DsCC achieves better accuracy and F1-score. The empirical study in this paper shows that the application of Mo-DsCC could be a useful puzzle for smart agriculture to benefit farms and bring new ideas to industry and academia.

\end{abstract}

\section{Introduction}
With the widespread employment of the Internet of Things (IoT) and machine learning (ML), Smart Agriculture (SA) has been growing rapidly in recent years. Farmers can operate and manage their farms at a large scale much easier than before. For example, ones can monitor, and track the state of crops or livestock from remote timely. That's not the most exciting part though, as with the recent development of machine learning technology, including Deep Learning (DL), we can leverage the prediction capacity to make further improvements for Smart Agriculture. Based on its power to learn from large data, DL has already shown huge potential in various agriculture tasks, including but not limited to fruit detection \cite{LU2020105760}, weed control \cite{2b1bee20d14245e49c23b3dfe45cded2}, farm’s weather prediction \cite{edssjs.4D1044AE20210101} and of course, the main point of this article, plant species and disease classification \cite{S187705092030690620200101,9418013,S004579061930002320190601}.

Leaf disease is one of the major causes of reduced crop yields or even crop failure. Compared with other external conditions, such as floods and droughts and other uncontrollable uncertain factors, early identification of plant diseases can greatly avoid the occurrence of foreseeable yield reduction. Among other parts of a plant, leaves are more visible and, therefore, leaf disease can be identified and diagnosed easier and sooner. Another feature that makes leaves popular in disease classification is that leaves participate in the photosynthesis process which will provide energy to the plant \cite{edssjs.15D0A2D220200101}. However, there are also challenges and difficulties for humans to classify leaf disease manually. It would take a substantial amount of labour, time and cost to identify the variety of leaves at scale. Most importantly, the annotating operators need professional knowledge and training time. Therefore, an efficient early identification approach of leaf diseases from various plant species is always the industry's desire, which can control the disease and treat the plants rapidly.

The issue above can be mitigated by the recent development of machine learning. In this context, machine learning algorithms are trained to recognize patterns and features in images of diseased and healthy leaves, so that they can accurately identify the types of disease in new images. This is done by using a dataset of labelled images, where the algorithm learns to differentiate between different types of diseased leaves based on the patterns and features present in the images. This can be a useful tool for farmers and plant researchers, as it can help to quickly and accurately identify diseases in crops, allowing for early intervention and treatment. Among different machine learning approaches, deep learning has shown great advantages and achieved state-of-the-art performance on multiple datasets \cite{S004579061930002320190601, 9155585, 9418245}. Another challenge in leaf disease detection is modelling the relationship between different plant species and disease types. While there are many different types of plants, each species may have unique characteristics when it comes to leaf diseases. Additionally, different species may have similar leaf diseases that can be treated in similar ways. Currently, most research in this field focuses on either classifying plant species or identifying specific diseases, rather than considering both factors simultaneously.

The purpose of this research is to explore an effective method of integrating plant species and leaf disease prediction. It is commonly observed that certain plants are susceptible to specific diseases, and similarly, certain diseases tend to affect specific plant species. This insight led us to the hypothesis that by considering both plant species and leaf diseases together we can reduce the number of false predictions. To this end, we propose a deep learning model, named Multi-output Deep Supervised Classifier Chains (Mo-DsCC), for plant identification and disease detection. Mo-DsCC is a multi-output deep convolutional neural network based on a modified VGG-16 backbone. We implement chaining of output layers on top of the backbone and employ deep supervision training for all output layers and the top layer of the backbone. In the experiment, we evaluate the proposed model on two benchmark leaf disease classification datasets. In particular, we use the Plant Village dataset for an evaluation in a laboratory environment and the PlantDoc dataset for an evaluation in a real-field environment. The results show that our model achieves improved performance than a wide range of models (AlexNet, VGG-16, ResNet101, InceptionV3, MobileNetV2, EfficientNet) and approaches including multi-model, power-set labelling, multi-output, and multi-task learning.

The organisation of our paper is as follows. In the next section, we discuss the related work. After that, in Section \ref{sec:Mo-DsCC} we describe our model, Multi-output Deep Supervised Classifier Chains. Section \ref{sec:exp} will present the experiments and showcase the effectiveness of our model. Finally, we conclude the paper and discuss our future work in Section \ref{sec:cf}.

\section{Related Work}
\label{sec:rw}

Shallow learning (SL), as the pioneer of the application of AI  in leaf species or disease classification, has achieved some success in the past \cite{154722790, 10.3389/fpls.2020.00751}. The advent of Deep Learning (DL) technology brought about a significant shift in the field. Many researchers have noted the superior effectiveness of Deep Learning (DL) in leaf image classification when compared to traditional machine learning methods \cite{SUJATHA2021103615, 9057889, 9418013}.

Convolutional Neural Networks (CNN), a famous class or a family of DL, has been widely used in plant species classification or disease classification. The CNN model family encompasses a variety of models with different characteristics. Some, like MobileNet \cite{mwebaze2019icassava, 9392051} and EfficientNet \cite{15107730820210601}, are designed for mobile deployment and have a relatively small computational footprint. Others, like AlexNet \cite{S004579061930002320190601}, GoogLeNet \cite{9418245}, VGG \cite{S187705092030690620200101, t_14770449620201201, 9231174}, and Inception \cite{S187705092030690620200101, 15100606920210401, electronics10121388}, require substantial computational resources to run. Other models such as ResNet \cite{t_14770449620201201} and MobileNet \cite{mwebaze2019icassava}, have unique characteristics that set them apart from the rest of the family.

Different from the previous approaches which are purely based on off-the-shelf CNNs, this paper introduces a multi-output model with output-layer chaining and deep supervision. 

\section{Multi-output Deep Supervised Classifier Chains}
\label{sec:Mo-DsCC}
\begin{figure*}[ht]
		\centering
		\includegraphics[width=0.6\textwidth]{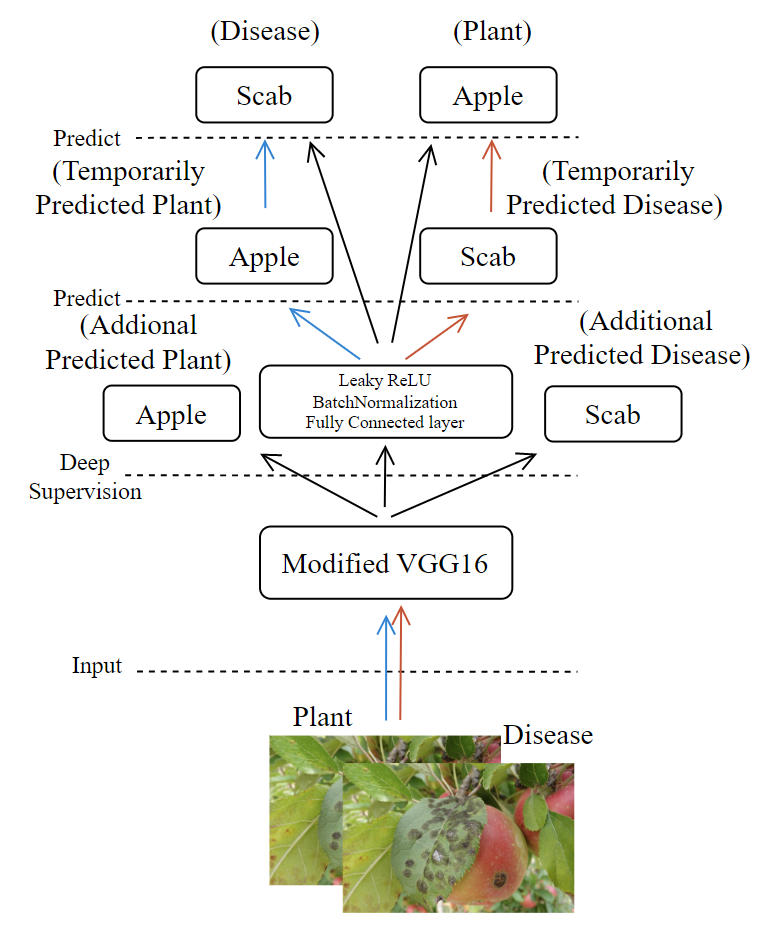}
		\caption{Chaining prediction layers of Modified VGG-16.}
		\label{fig:VGG}
	\end{figure*}

In this section, we propose Multi-output Deep Supervised Classifier Chains (Mo-DsCC), a deep network for plant identification and disease classification. Our model consists of three components: A modified VGG-16 network as the backbone, a stack of classification chains and deep supervision training. The architecture of our Mo-DsCC is shown in Figure \ref{fig:VGG}.

\subsection{Modified VGG-16}
In Figure \ref{fig:VGG}, Mo-DsCC is built on top of a convolutional neural network. This backbone CNN is a variant of VGG-16. We update the architecture of VGG-16 to improve the prediction of multiple outputs in this particular application of plant identification and disease classification. Based on the original VGG-16 model, we did some modifications to improve the performance of leaf species and disease classification. The original VGG-16 shone through in the 2014 ImageNet challenge, its top-5 test accuracy archived at $92.7\%$. Its structure contains 13 convolutional layers and their kernel filters are $3\times 3$, 5 max-pooling layers, and 3 fully connected layers. The max pooling layer split the networks into 6 blocks. These 6 blocks are 2, 2, 3, 3, 3 convolutional layers and 3 fully connected layers. The size of the second and third fully connected layers from the bottom is $4,096$ nodes, and the bottom layer’s units will be configured based on the class number of a classification task. To further enhance performance, we have implemented batch normalization after each convolutional and fully connected layer and utilized Leaky ReLU as the activation function. Additionally, we have replaced the Flatten layer with a Global Average Pooling (2D) layer between the last max pooling layer and the first fully connected layer.

\subsection{Chaining prediction layers}
\label{chaining}
On top of the backbone CNN (our modified VGG-16), we create a stack of output layers. Our stacking idea is inspired by the "classifier chains" in \cite{Read_2009}. However, different from the original idea of chaining of classifiers in \cite{Read_2009}, we chain the prediction layer in our deep networks so that the prediction of one label can be employed to predict the others. Firstly, the modified VGG-16 will predict two Temporarily Predicted outputs (we call them: $Plant\_t \ \& \ Disease\_t$) through the image’s features ($x$). Then, alongside the Plant\_t, it will with the image’s features ($x$) predict the final Disease label of the leaf image (i.e., 	$Plant\_t + x \Rightarrow Disease$). Vice versa, the final Plant will be predicted through Disease\_t and image features, (i.e., $Disease\_t + x \Rightarrow Plant$). The theory is based on the assumption that certain diseases exist only in certain plants, and certain plants contain only certain diseases. Therefore, by sharing some features, these two labels could eliminate distracting options for each other.

\subsection{Deep Supervision}
\label{DS}
After the first fully-connected layer of the modified VGG-16, we deployed the deep supervision method. Deep supervision, its other names may be Auxiliary Supervision or Intermediate Supervision, is a method to add additional classifiers or computing loss in the hidden layers of the deep neural network, which can provide additional supervision to improve the final label prediction performance \cite{Wang2015TrainingDC, 02376}. For example, in the intermediate layers of the proposed model of \cite{7298594}, they added additional classifiers which are smaller networks for additional supervision to improve the training performance.
However, the types and placement of classifiers have not been determined yet and need to be analysed on a task-specific basis. For example, \cite{pmlr-v38-lee15a} advocated using SVMs with a squared hinge loss as the classifiers, but \cite{Wang2015TrainingDC, 7298594} proposed using small neural networks. \cite{02376} analysed three different positions in the neural networks, i.e., shallow layers, intermediate layers and deep layers, which have their own different advantages and disadvantages and should be taken task-specific. In this study, we added two fully-connected layers with SoftMax function after the Global Average Pooling 2D layer after our modified VGG-16 model and before the fully-connected layer and the chaining prediction layers after that, to provide additional supervision for two target labels (i.e., species \& diseases).

\subsection{Training}
\label{Training}
As mentioned earlier, our approach (Mo-DsCC) is inspired by the relationship between plant species and diseases, i.e., the two labels will benefit each other in training. Therefore, we added weights for these different losses of output (SoftMax) layers at different levels. In Figure \ref{fig:VGG}, we can see three special features, (1), Deep Supervision (DS) layers (2),  different tasks have separate dense layers. (3), these prediction layers are stacking. Firstly, the convolutional layers (in the modified VGG-16) will learn the global features of the leaf images and these separate dense layers will learn the specific features of each task (i.e., plant or disease). Deep Supervision (DS) layers will first predict two additional outputs in the first prediction level, named here as $p^{a}$ (i.e., Plant Additional) and $d^{a}$ (i.e., Disease Additional) to provide additional supervision. The second prediction layer will predict the temporary plant and disease results (i.e., $p^{temp}$ \& $d^{temp}$). After that, the cross-connection method will be applied and it will concatenate the probability of $p^{temp}$ or $d^{temp}$ with the global features to predict the final predicted plant species ($p$) and leaf diseases ($d$) in the third prediction level. All outputs (i.e., $p^{a}$, $d^{a}$, $p^{temp}$, $d^{temp}$, $p$ \& $d$) will be recorded to show our approach is effective to improve the final prediction results ($p$  \& $d$).

As mentioned earlier, there are 6 outputs of Mo-DsCC (i.e., $p^{a}$, $d^{a}$, $p^{temp}$, $d^{temp}$, $p$ \& $d$) and each loss should be minimised. In order to adjust the relationships and constraints among each output and improve the performance of the final two outputs ($p$ \& $d$), we set a weight for each output and train Mo-DsCC by minimising the following loss function:

\vskip -0.2cm
\begin{equation}
\begin{split}	
	\mathcal{C} = \beta_a \mathcal{L}(P_a,\hat{P}_a) +\delta_a \mathcal{L}(D_a,\hat{D}_a) + \\ 
	\beta_1 \mathcal{L}(P_1,\hat{P}_1) + \beta_2 \mathcal{L}(P_2,\hat{P}_2) + \\
	\delta_1 \mathcal{L}(D_1,\hat{D}_1) + \delta_2 \mathcal{L}(D_2,\hat{D}_2)
\end{split}	
\end{equation}

In the formula, $\mathcal{L}$ is the cross-entropy loss function;  $P_a$, $D_a$ are the ground truth of additional plant species and disease at the first prediction level (level 1) to provide additional supervision, $P_1$, $D_1$ are the ground truth of temporary plant species and disease at the second prediction level (level 2), and $P_2$, $D_2$ are the ground truth for final plant species and disease at the third prediction level (level 3) which are the final predicted results of plant and disease of leaf images. $P_1$, $D_1$ will temporarily infer the probability of the final $D_2$, $P_2$. $\beta_a$, $\delta_a$, $\beta_1$, $\beta_2$, $\delta_1$, $\delta_2$ are the balance weights and the hyper-parameters of this study. We fine-tune these balance weights through grid-search for better training performance and try to understand why and how they interact. We expect these weights to have an impact on the final prediction.

\subsection{Inference}
There are three stages of inference for plant types and disease types in our approach (Mo-DsCC). All outcomes of these three prediction layers will be recorded. In general, only the final prediction layers (i.e., the $3^{rd}$ layer) will be the final result of Mo-DsCC. The $1^{st}$ layer is the deep supervision layer and the $2^{nd}$ layer is the temporary prediction layer. However, changing the balance weights of each output will cause the structure and usage of Mo-DsCC to change, which will make the model’s outputs close to multi-model and multi-output approaches, as special cases.

\begin{itemize}
\item  When $\beta_1=1$ and $\beta_2=\delta_1 = \delta_2 = 0$, in this situation, Mo-DsCC is similar to a single CNN model to predict plant species. $P_1$ in the first level will be the only final output of Mo-DsCC.
\item   When $\delta_1=1$ and $\beta_1=\beta_2 = \delta_2 = 0$, in this situation, Mo-DsCC is similar to a single CNN model to predict disease types. $D_1$ in the first level will be the only final output of Mo-DsCC.
\item When $\beta_1=\delta_1=1$ and $\beta_2$ = $\delta_2$=0, in this situation, Mo-DsCC is similar to a multi-output CNNs. Both $P_1$ and $D_1$ in the first level will be the final outputs of Mo-DsCC.
\end{itemize}

\section{Experiments}
\label{sec:exp}
\subsection{Setting}
This study’s experiments are based on Python 3.7, the scripts of Multi-model, Multi-label, Multi-output, Cross-stitch Network \cite{Misra_2016_CVPR} of Multi-task, and our proposed approach Mo-DsCC are based on Tensorflow-2.6, and other 3 Multi-task approaches (MTAN \cite{Liu_2019_CVPR}, TSNs \cite{10.1007/978-3-319-24574-4_28} \& MOON \cite{Li_2021_CVPR}) are based on Pytorch-1.11.0. The datasets are Plant Village and PlantDoc (See Section \ref{datasets}). The ratio of the training set and test set of Plant Village is 80:20 and PlantDoc has its own test set. 10\% of the training set will be split as the validation set. The input size of leaf images was set as $256 \times 256$ without data augmentation and their colour model is RGB. The early stopping method has been adopted to monitor the validation loss of model training to avoid the over-fitting issue, it will stop training when it finds that the loss function doesn't decrease after 50 epochs and the max epoch number is 10000. For the hyper-parameters, the learning rate is 0.001, the batch size is 16, and for the optimizer, we selected ‘Adamax’. Each approach will be run 10 times to verify the results’ reliability, both accuracy and F1-score of times will be calculated as the average with standard deviation.

\subsection{Datasets} 
\label{datasets}
	We selected two public datasets in this experiment (see Table \ref{Public_Dataset_Address}).
	\begin{table*}[h!]
		\begin{center}
			\begin{minipage}{0.8\textwidth}
				\caption{Datasets of This Study}\label{Public_Dataset_Address}%
				\resizebox{\textwidth}{!}{%
					\begin{tabular}{@{}llccccl@{}}
						\hline
						ID & Dataset & Year & Species & Category & Class & Link \\
						\hline
						1 & Plant Village \cite{hughes2016open}  & 2016 & 14 & 22 & 38 & \url{https://data.mendeley.com/datasets/tywbtsjrjv/1}\\
						2 & PlantDoc \cite{Singh_2020} & 2020 & 13 & 17 &28 & \url{https://github.com/pratikkayal/PlantDoc-Dataset} \\
						\hline
					\end{tabular} %
				}
			\end{minipage}
		\end{center}
	\end{table*}
	
	\subsubsection{Plant Village Dataset}
	The Plant Village dataset is widely recognized as a prominent benchmark in plant leaf disease classification tasks. It was released in 2016 \cite{hughes2016open}, and contains $54,305$ healthy or disease leaf images including 14 species of plants (i.e., (Apple, Blueberry, Cherry, Corn, Grape, Orange, Peach, Bell Pepper, Potato, Raspberry, Soybean, Squash, Strawberry \& Tomato). Each plant contains a category of different states (e.g., healthy, scab or late blight), among them, the tomato has the most categories (10). Totally, there are 38 categories of joint-species-disease labels (e.g., Apple Black Rot). The author released with data augmentation version in 2019 \cite{DBLP:journals/corr/HughesS15}, the original dataset has been increased from $54,305$ to $61,486$ leaf images through 6 augmentation methods. We used the original version in this study. We split the data into $80\%$-$20\%$ for training and test sets respectively, and further, split $10\%$ from the training set as the validation set. 
	
	\subsubsection{PlantDoc Dataset}
	
	The idea of PlantDoc Dataset was created based on Plant Village. The author thought the images of Plant Village are all from the laboratory and they are not the leaf images of a real field, thus, they collected the lea images from the cultivation field \cite{Singh_2020}, and tried to make the PlantDoc a challenging benchmark dataset. The PlantDoc has $2,598$ leaf images, 13 species, 17 unique disease states, and 28 categories of joint-species-disease labels. It has its original training set and test set, we split $10\%$ from the training set as the validation set.

\subsection{Evaluation Metrics}

Accuracy and F1-score may be the most two suitable evaluation metrics for classification tasks. Their calculations can be based on confusion matrix \cite{edsdoj.8f0d61eb104a67a1e887ad1111fbd220200101}. Accuracy can be can initially provide analysts with the prediction efficiency of a machine learning or deep learning model. It means the ratio of the number of the right prediction results to all prediction results in one classification task Formula (See Formula \ref{ACC}. For example, suppose we need to predict 100 cats and dogs. In the results, 30 cats and 40 dogs are accurate, so the accuracy is (30+40)/100 = 70\%. However, Accuracy has its limitation in that it will not reflect the performance of the model well when facing imbalanced data prediction tasks.  The most famous example is the spam email prediction task, in total 100 emails have 2 spam, it is hard to claim a model which accurately predicts 70\% of normal mail is an efficient model. Thus, using only Accuracy cannot fully evaluate the predictive performance of the model. F1-score will be more useful when facing imbalanced data tasks. F1-score is calculated as Formula \ref{f1_score}:
	\begin{equation}
		\textit{$Sensitivity (Recall) = \frac{TP}{(TP+FN)}$}
	\end{equation}
	\begin{equation}
		\textit{$Precision = \frac{TP}{(TP+FP)}$}
	\end{equation}
	\begin{equation}
		\textit{$Accuracy = \frac{(TP+TN)}{(TP+FN+FP+TN)}$}
		\label{ACC}
	\end{equation}
	\begin{equation}
		\textit{$F1$-$score = \frac{(2 \ast Precision\ast  Recall)}{(Precision+ Recall)}$}
		\label{f1_score}
	\end{equation}

In this study, both the Accuracy and F1-score of three classification tasks will be calculated, i.e., plant species, leaf disease and joint-species-disease label. In the joint-species-disease label, both plant species and leaf disease of leaf images should be accurate.

\subsection{Competitors}

To address the challenge of simultaneously predicting plant species and diseases, an appropriate approach is very important. Therefore, we have explored and tested several different approaches and proposed our improvement approach, i.e., Multi-output Deep Supervised Classifier Chains (Mo-DsCC). This paper aims to use an efficient single model to predict both species and diseases simultaneously. The DL approaches we researched are as follows:

\begin{itemize}
	\item Multi-model: This is the most basic DL method which uses several single models to solve several production tasks. In this case, i.e., using one baseline model to predict the species of the leaf images and another one to predict the diseases of the leaf images. 
	
	\item Multi-label:  This approach will use a single baseline model to predict both two labels through the Power Set method which means combining two labels into one label, it contains both species and disease information. This model accepts one input and one output.

	\item Multi-output: This is also a single-model approach, different to Multi-label, this approach uses one input but produces several outputs to handle multiple tasks simultaneously. In this case, it will use one baseline model to produce two predicted result outputs which are species and diseases. 
	
	\item Multi-task. Multi-task learning has been taken to solve this problem, in this task, there are two different target tasks, but we will use the same leaf images as input data. Combine with DL, Multi-task learning becomes more efficient in computer vision, because different tasks will benefit from each other’s learning \cite{9392366, 10.1007/978-3-319-10599-4_7}. In this part, we selected and tested 4 recent Multi-task approaches, i.e., Cross-stitch Network \cite{Misra_2016_CVPR}, Multi-Task Attention Network (MTAN) \cite{Liu_2019_CVPR}, Task Switching Networks (TSNs) \cite{10.1007/978-3-319-24574-4_28} and Model-Contrastive Learning (MOON) \cite{Li_2021_CVPR}.
\end{itemize}

In our evaluations of multi-model, Multi-label and multi-output approaches, we experimented with various CNN backbones, such as AlexNet, VGG-16, ResNet101, InceptionV3, MobileNetV2, and EfficientNet. Out of these, we found that InceptionV3 delivered the best results and will use it as the benchmark for comparison.

\subsection{Experimental Results}

Table \ref{tab:plant_village_leaves} presents the experimental results of this study, which are divided into three categories: Plant (plant species), Disease (leaf disease), and Total (joint species-disease label). The Plant category reflects the accuracy of the model in predicting the species of the target leaf, the Disease category demonstrates the model's ability to correctly identify the type of disease present in the target leaf image, and the Total category assesses the model's overall performance in correctly identifying both the plant species and disease of a given leaf. Both accuracy and F1-score are used as evaluation metrics, and the results have been obtained by running the experiment 10 times and reporting the mean and standard deviation of the results.

The last row in Table \ref{tab:plant_village_leaves} shows the accuracy and F1-score of our proposed model (Mo-DsCC). As mentioned above, the balance weights of Mo-DsCC are searched by the grid-search method, and the locally optimal hyper-meters so far are selected as $\beta_a:\delta_a:\beta_1:\delta_1:\beta_2:\delta_2=0.1:0.1:0.1:0.1:0.4:0.5$.

Firstly, the results of the Plant Village dataset indicate that models are able to learn effectively from this dataset, as most of the results are above 0.90. Among the deep learning (DL) approaches, Multi-label performed the best in the Plant category, with an accuracy of 0.99810 and an F1-score of 0.99810. In the Disease category, Multi-output outperformed traditional and recent Multi-task approaches with an accuracy of 0.99413 and an F1-score of 0.99413. However, the Total category, which combines the accuracy of the Plant and Disease labels, resulted in lower scores. In the Total task of Plant village, Multi-label achieved the highest accuracy of 0.99310 and an F1-score of 0.99311. The results of our proposed model, Mo-DsCC, are even more impressive. In the Plant category, Mo-DsCC obtained an accuracy of 0.99911 and an F1-score of 0.99911, which is even better than Multi-label. In the Disease category, Mo-DsCC achieved an accuracy of 0.99773 and an F1-score of 0.99773, surpassing Multi-output. Finally, in the Total category, Mo-DsCC achieved an accuracy of 0.99705 and an F1-score of 0.99718, overwhelmingly outperforming Multi-label.

Secondly, in the PlantDoc dataset, the results also demonstrate a clear improvement with our proposed model, Mo-DsCC. Among traditional DL approaches and recent Multi-task methods, Multi-output performed best in the Plant category, with an accuracy of 0.43093 and an F1-score of 0.38037. In the Disease category, Multi-model achieved the highest performance, with an accuracy of 0.51992 and an F1-score of 0.48344. In the Total task, Multi-label outperformed the others, with an accuracy of 0.25466 and an F1-Score of 0.21130. However, our Mo-DsCC method achieved even better results in the Plant category, with an accuracy of 0.55000 and an F1-score of 0.53036, surpassing Multi-output by a significant margin. In the Disease category, Mo-DsCC achieved an accuracy of 0.60678 and an F1-score of 0.58586, outperforming Multi-model. Finally, in the Total task, Mo-DsCC achieved an accuracy of 0.34364 and an F1-score of 0.32501, showing an improvement of around 10\% compared to the previous winner, Multi-label.

It is evident from the results that our proposed method Mo-DsCC has effectively improved the performance of both plant and disease prediction. This supports our hypothesis that the correlation between plant species and leaf diseases can be utilized to reduce false predictions.
\begin{table*}[ht]
	\centering
	{\scriptsize
		\begin{tabular}{|l|c|c|c|c|c|c||c|c|c|c|c|c|}
			\hline
			& \multicolumn{6}{|c||}{Plant Village}& \multicolumn{6}{|c|}{PlantDoc}\\
			Models & \multicolumn{2}{|c|}{Plant}& \multicolumn{2}{|c|}{Disease}& \multicolumn{2}{|c||}{Total}& \multicolumn{2}{|c|}{Plant}& \multicolumn{2}{|c|}{Disease}& \multicolumn{2}{|c|}{Total}\\
			& Acc. & F1 & Acc & F1& Acc. & F1& Acc. & F1 & Acc & F1& Acc. & F1 \\
			\hline
			
			Multi-model & $\underset{\pm 0.00074}{0.99389}$ & $\underset{\pm 0.00075}{0.99388} $ 
			& $\underset{\pm0.00074}{0.98561} $	&$\underset{\pm0.00076}{0.98558}$
			& $\underset{\pm0.00000}{0.98034} $	&$\underset{\pm0.00029}{0.98301} $   
			& $\underset{\pm0.03301}{0.40297} $ &$\underset{\pm0.03833}{0.37522} $ 
			& $\underset{\pm0.03272}{0.51992} $	&$\underset{\pm0.05286}{0.48344} $ 
			& $\underset{\pm0.02924}{0.20212} $	&$\underset{\pm0.04271}{0.19346}$ \\
			\hline
			Multi-label & $\underset{\pm0.00118}{0.99810} $ &	$\underset{\pm0.00118}{0.99810} $ & $\underset{\pm0.00099}{0.99383} $	&$\underset{\pm0.00099}{0.99384} $& $\underset{\pm0.00171}{0.99310} $	&$\underset{\pm0.00173}{0.99311} $&  $\underset{\pm0.00636}{0.38771} $   &$\underset{\pm0.00783}{0.36693} $& $\underset{\pm0.00381}{0.46483} $	&$\underset{\pm0.00194}{0.43180} $& $\underset{\pm0.00127}{0.25466} $	&$\underset{\pm0.00654}{0.21130} $ \\
			\hline
			Multi-output & $ \underset{\pm 0.00044}{0.99752} $ &$ \underset{\pm 0.00044}{0.99753}$ 
			& $ \underset{\pm 0.00056}{0.99414} $ &$\underset{\pm0.00056}{0.99413} $
			& $ \underset{\pm 0.00017}{0.99215} $ &$\underset{\pm0.00018}{0.99239} $ 
			& $ \underset{\pm 0.03902}{0.43093} $ &$\underset{\pm0.04974}{0.38037} $ 
			& $ \underset{\pm 0.03239}{0.46441} $ &$\underset{\pm0.03922}{0.40202} $ 
			& $ \underset{\pm 0.04308}{0.18136} $ &$\underset{\pm0.03692}{0.14635} $\\
			\hline
			Cross-stitch & $\underset{\pm 0.00155}{0.98197}$& $\underset{\pm 0.00153}{0.98196}$ 
			& $\underset{\pm 0.00169}{0.95241}$ & $\underset{\pm 0.00179}{0.95237}$
			& $\underset{\pm 0.00310}{0.93870}$ & $\underset{\pm 0.00276}{0.94509}$ 
			& $\underset{\pm 0.02734}{0.38814}$ & $\underset{\pm 0.03165}{0.34577}$ 
			& $\underset{\pm 0.01431}{0.44492}$ & $\underset{\pm 0.01102}{0.38920}$
			& $\underset{\pm 0.02241}{0.16610}$ & $\underset{\pm 0.01913}{0.15028}$\\
			\hline
			MTAN &$\underset{\pm 0.00390}{0.95461}$& $\underset{\pm 0.00395}{0.95464}$
			& $\underset{\pm 0.00710}{0.91410}$ & $\underset{\pm 0.00702}{0.91411}$
			& $\underset{\pm 0.00898}{0.89509}$ & $\underset{\pm 0.00836}{0.90269}$ 
			& $\underset{\pm 0.01180}{0.30169}$ & $\underset{\pm 0.01435}{0.18824}$ 
			& $\underset{\pm 0.01079}{0.38941}$ & $\underset{\pm 0.01672}{0.25224}$
			& $\underset{\pm 0.00988}{0.06102}$ & $\underset{\pm 0.00839}{0.02871}$\\
			\hline
			TSNs & $\underset{\pm 0.01452}{0.91704}$ & $\underset{\pm 0.01468}{0.91686}$   
			& $\underset{\pm 0.02681}{0.79926}$ & $\underset{\pm 0.02653}{0.79900}$
			& $\underset{\pm 0.03190}{0.76808}$ & $\underset{\pm 0.03101}{0.77433}$ 
			& $\underset{\pm 0.01989}{0.32881}$ & $\underset{\pm 0.03381}{0.21219}$ 
			& $\underset{\pm 0.01254}{0.37627}$ & $\underset{\pm 0.02069}{0.25399}$
			& $\underset{\pm 0.01545}{0.05975}$ & $\underset{\pm 0.01049}{0.02682}$\\
			\hline
			MOON & $\underset{\pm  0.00410}{ 0.97589}$ &  $\underset{\pm  0.00411}{ 0.97586}$  
			& $\underset{\pm  0.00889}{ 0.95307}$ & $\underset{\pm  0.00886}{ 0.95296}$
			& $\underset{\pm  0.00997}{ 0.94414}$ & $\underset{\pm  0.00974}{ 0.94586}$ 
			& $\underset{\pm  0.01542}{ 0.35339}$ & $\underset{\pm  0.01692}{ 0.25096}$ 
			& $\underset{\pm  0.01485}{ 0.41568}$ & $\underset{\pm  0.02726}{ 0.31160}$
			& $\underset{\pm  0.01506}{ 0.09068}$ & $\underset{\pm  0.01521}{ 0.04485}$\\
			\hline
			\hline	
			Mo-DsCC & $\underset{\pm 0.00016}{0.99911}$&  $\underset{\pm 0.00016}{0.99911}$ 
			& $\underset{\pm 0.00049}{0.99773}$ &  $\underset{0.00049\pm}{0.99773}$
			& $\underset{\pm 0.00057}{0.99705}$ &  $\underset{\pm 0.00057}{0.99718}$ 
			
			& $\underset{\pm 0.03873}{0.55000}$ &  $\underset{\pm 0.03192 }{0.53036}$ 
			& $\underset{\pm 0.02403}{0.60678}$ &  $\underset{\pm 0.02630 }{0.58586}$
			& $\underset{\pm 0.04466}{0.34364}$ &  $\underset{\pm 0.04244 }{0.32501}$ \\
			\hline
		\end{tabular}
	}
	\caption{The Experimental Results (Plant Village \& PlantDoc)}

	\label{tab:plant_village_leaves}
\end{table*}

\subsection{Results of Additional and Temporary Outputs}


\begin{figure}[h!]
	\begin{center}
			\begin{minipage}{0.5\textwidth}
				\includegraphics[width=0.9\textwidth]{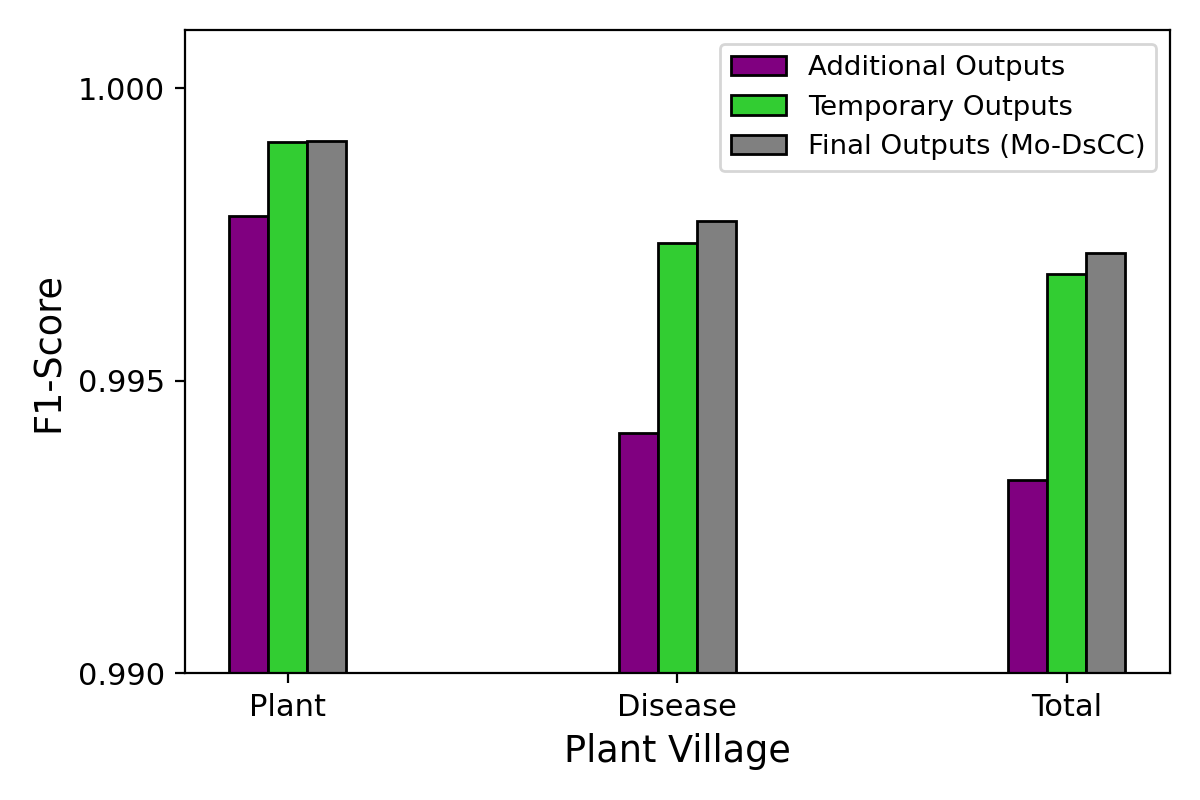}
				\label{fig:PlantVillage_add_temp}
			\end{minipage}
			\begin{minipage}{0.5\textwidth}
				\includegraphics[width=0.9\textwidth]{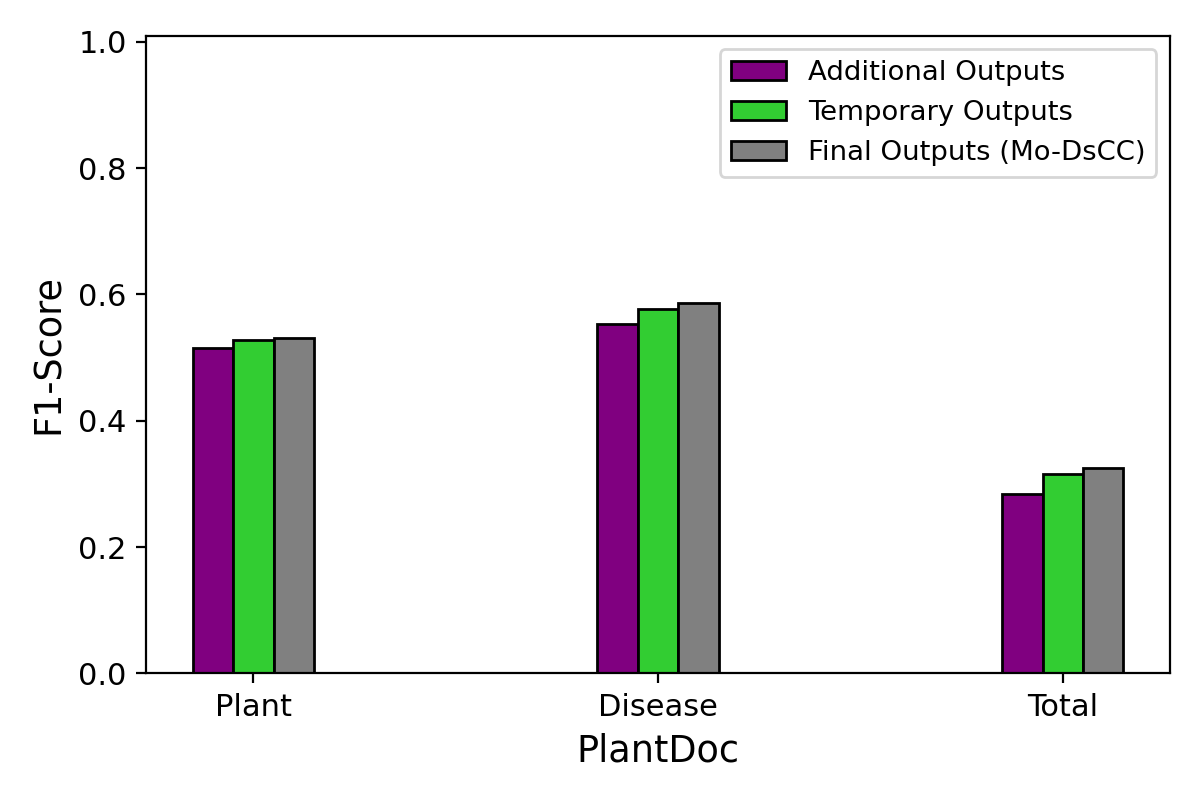}
				\label{fig:PlantDoc_add_temp}
			\end{minipage}
		\caption{Results of Additional and Temporary Outputs.}
		\label{fig:addtemp}
	\end{center}
\end{figure}

As depicted in Section \ref{Training} and Figure \ref{fig:VGG}, $p$ \& $d$ are the final outputs of the model (i.e.,), $p^{a}$ and $d^{a}$ are derived from the outputs of the Deep Supervision (DS) layers, and $p^{temp}$ \& $d^{temp}$ are the temporary plant and disease results from the second prediction layer. Figure \ref{fig:addtemp} illustrates the relationship among the Additional, Temporary and Final layers of Mo-DsCC. As the model progresses from the input to the final output layer, the F1-scores for the three tasks of Plant, Disease, and Total all increase, demonstrating that the performance of the model has been improved through the incorporation of deep supervision and classifier chains.

\subsection{Effect of Classifier Chains}

\begin{figure}[h!]
	\begin{center}
			\begin{minipage}{0.5\textwidth}
				\includegraphics[width=0.9\textwidth]{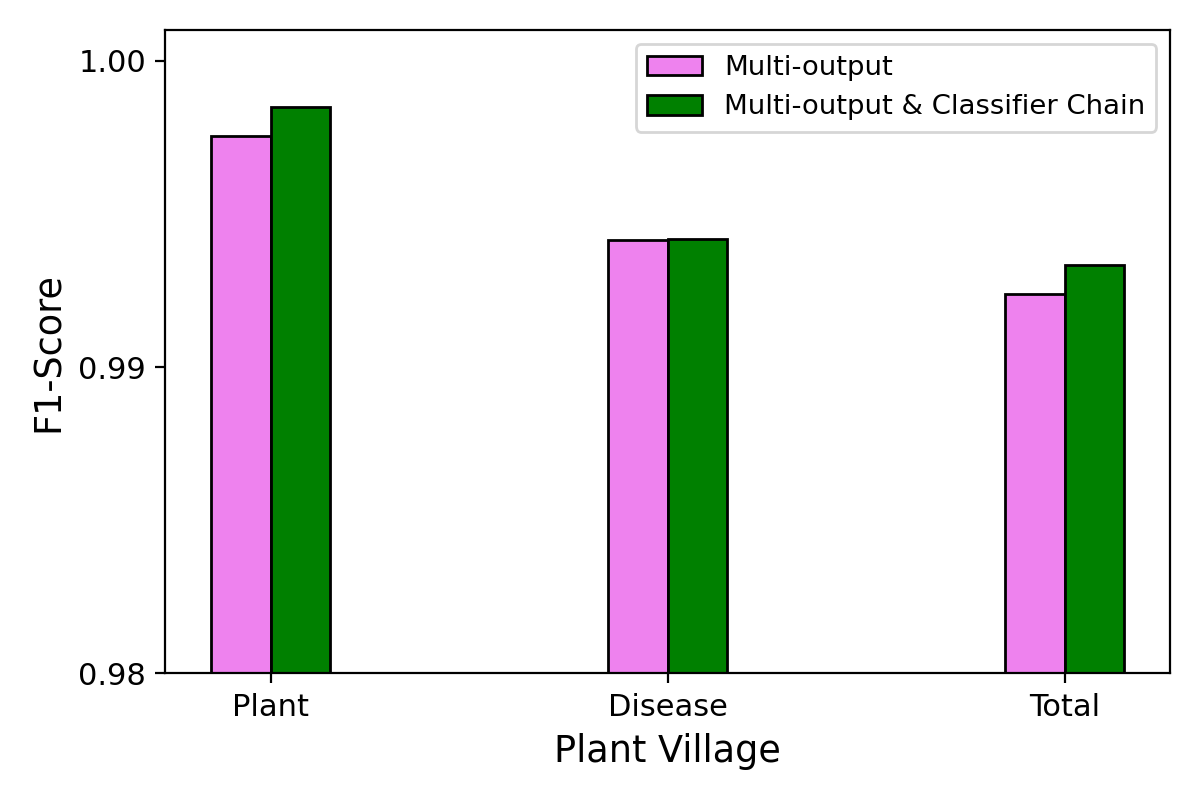}
				\label{fig:wcc_PlantVillage_wcc}
			\end{minipage}
			\begin{minipage}{0.5\textwidth}
				\includegraphics[width=0.9\textwidth]{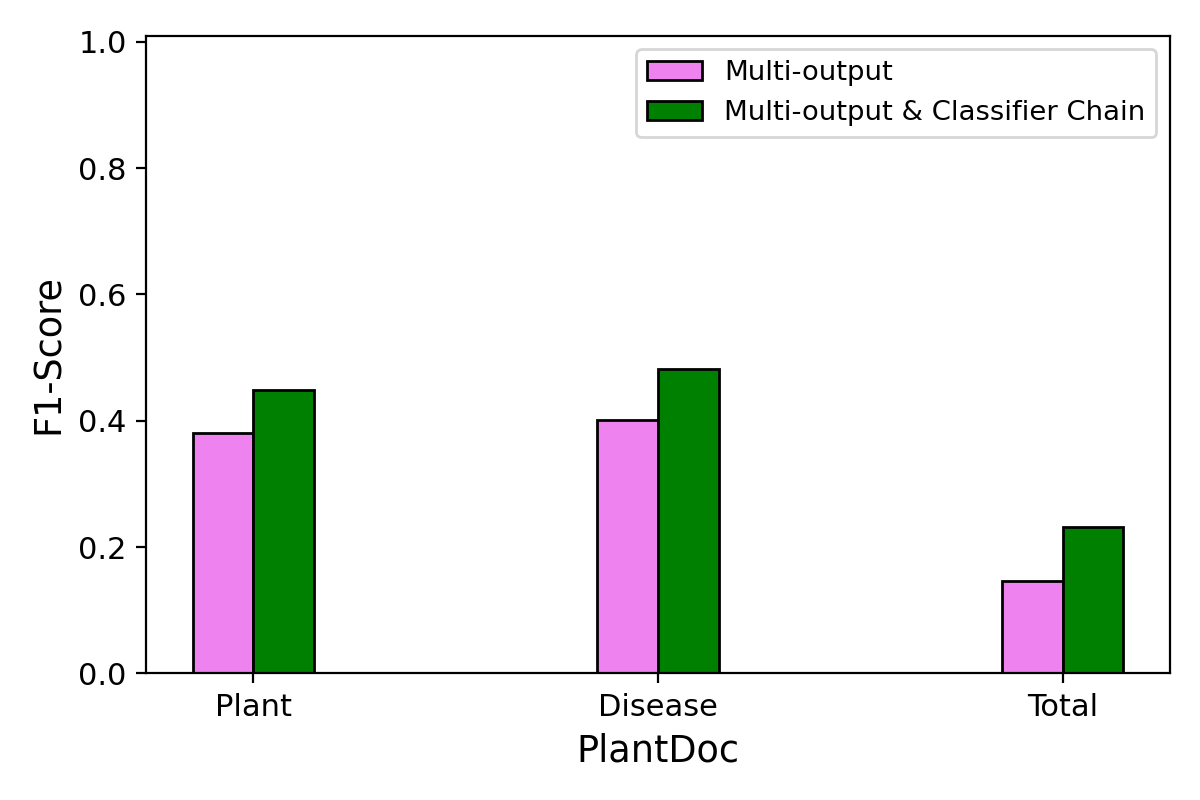}
				\label{fig:wcc_PlantDoc_wcc}
			\end{minipage}
		\caption{Effect of Classifier Chains.}
		\label{fig:eocc}
	\end{center}
\end{figure}

Figure \ref{fig:eocc} illustrates the comparison between the use and non-use of the Classifier Chains method. The backbones used in both cases are Inception V3, represented by the green and pink histograms respectively. It is evident that in all 3 tasks and across 2 datasets, the performance of the model incorporating the Classifier Chains method (green histograms) has seen a significant improvement when compared to the model without it (pink histograms).

\subsection{Effect of Deep Supervision}

\begin{figure}[h!]
	\begin{center}
			\begin{minipage}{0.5\textwidth}
5				\includegraphics[width=0.87\textwidth]{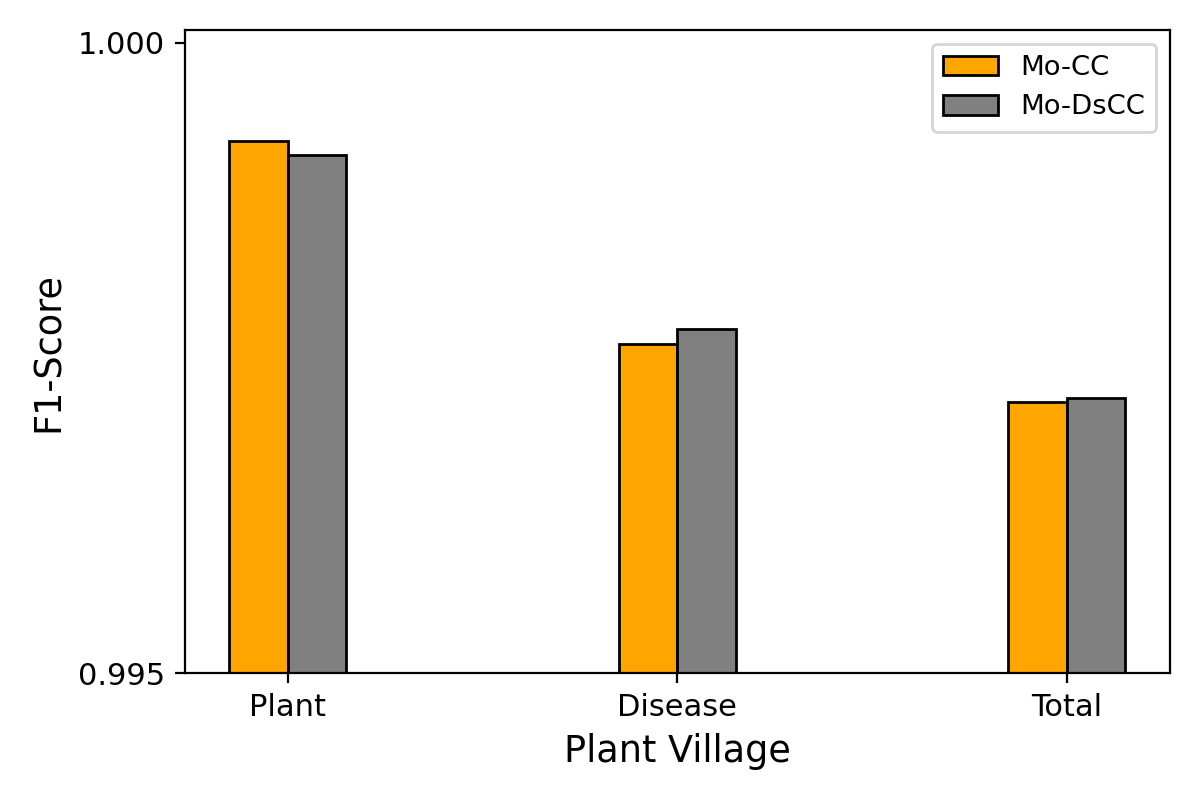}
				\label{fig:eds_PlantVillage_wcc}
			\end{minipage}
			\begin{minipage}{0.5\textwidth}
				\includegraphics[width=0.9\textwidth]{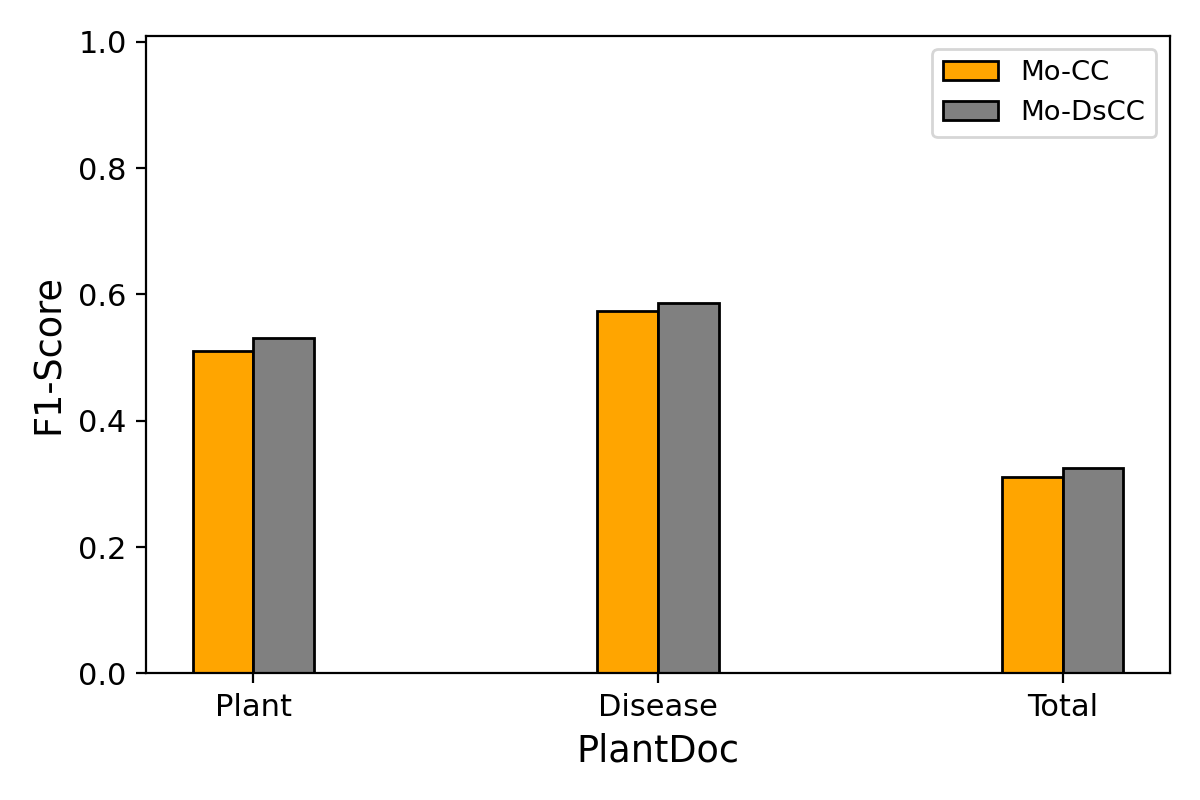}
				\label{fig:eds_PlantDoc_wcc}
			\end{minipage}
		\caption{Effect of Deep Supervision}
		\label{fig:eods}
	\end{center}
\end{figure}

Figure \ref{fig:eods} illustrates the impact of the Deep Supervision technique used in this study by comparing the performance of the Mo-DsCC model with and without it. The backbone used is a modified VGG-16. The orange histograms represent the Multi-output only with the Classifier Chains method (Mo-CC), while the grey histograms represent our proposed method (Mo-DsCC). As observed, the Mo-DsCC model only shows a slight performance decrease in the plant classification task of the Plant Village dataset. However, it shows improvement in the Disease task, and an overall improvement in the Total F1-score. In the PlantDoc dataset, the Mo-DsCC model shows improved performance in all three tasks.

\section{Conclusion and Future Work}
\label{sec:cf}
In this study, we propose a new approach called Multi-output Deep Supervised Classifier Chains (Mo-DsCC) for classifying plant species and diseases simultaneously or separately. This approach is beneficial for farms and industries that rely on traditional agricultural techniques for disease detection and diagnosis. Mo-DsCC is composed of a Modified VGG-16 model, Deep Supervision, and Chaining prediction layers parts, and it has been proven to be superior to other existing techniques and approaches, such as Multi-model, Multi-label, Multi-output, and recent Multi-task approaches, as seen in Table \ref{tab:plant_village_leaves}. The experimental results of Mo-DsCC are more accurate and reliable. We hope that our research will bring new ideas to both academia and industry, and accelerate the implementation and advancement of Smart Agriculture. In future work, we plan to expand the concept of deep-supervised classifier chains to general multi-output prediction tasks.

\bibliography{IEEEabrv,mybibfile}{}
\bibliographystyle{IEEEtran}

\end{document}